# Revisiting the syntax of imperatives in Yemeni Arabic: An Agree across phases approach

Mohammed Q. Shormani
Ibb University, Yemen
shormani@ibbuniv.edu.ye/https://orcid.org/0000-0002-0138-4793

**Abstract**

This article revisits the syntax of imperatives in Yemeni Arabic proposing an Agree Across Phases (AAP) approach. I argue that the AAP approach successfully accounts for both simple and complex imperative constructions, including A′-chain structures, by establishing a close interaction between syntax and discourse. The study demonstrates that this interface is motivated by the interpretive and performative functions associated with imperatives, linking informational structure with propositional structure. It is also proposed that the thematic subject of imperatives is a 2-person pro, whereas any overt pronominal or nominal element occurring preverbally is not a subject, but rather a C-domain element, precisely aboutness topic. These topics serve as the logical subjects of imperatives and enter into a coreferentiality relationship with pro. This relation is analyzed as AAP involving Match, yielding both local and non-local A′-chains. For core imperatives, viz., lacking an overt topic, I propose a null topic to (re)merge in Spec,TopP, whose interpretation depends on the discourse.

## 1. Introduction

Imperatives have been extensively studied cross-linguistically, and various approaches have emerged. This article aims to provide a novel approach to the analysis of imperatives, based on their syntactic and discourse properties and functions. It involves data from Yemeni Arabic (YA), and hopes to have a cross-linguistic appeal. It will be limited to imperative structures of the type in (1) (see Shormani 2021a & b).

(1) a. You stop talking!
b. Stop talking!
c. Boys leave!
d. Ali take the eraser and (*Ali) erase the board!
e. Boys leave and girls stay!

The existing analyses hold that the subject of an imperative clause can be an overt pronoun (1a), a covert pronoun (1b), or a lexical NP (1c). However, the question is: are *you* in (1a) and *boys* in (1c) subjects in the technical sense? If so, then another question arises: why are such constituents allowed to be dropped in virtually all languages?

Syntactically, it has long been recognized that the natural and normal form of imperatives is (1b) in virtually all languages, i.e., imperatives are subjectless constructions (see e.g., Downing 1969, Beukema and Coopmans, 1989, Han 1998, Jensen 2003, Rupp 2003, Bennis 2006, Van der Wurff 2007, Radford 2009, Kaufmann 2012, Alcázar and Saltarelli 2014). However, given the clausal nature of imperatives, they must have a subject in order to satisfy the EPP feature. If the overt (pro)nominal constituents in (1) are topics, then, the subject must be a covert/null pronoun. If this subject is null, then, there must be a mechanism under which it is licensed and interpreted. The issue of how a null pronoun is licensed in imperative structures has been extensively researched. In the literature, two types of null pronouns are assumed as the subjects of imperatives, namely PRO (see e.g., Han 1998), or *pro* (see e.g., Jensen 2003, Bennis 2006). As for *pro*, two salient views can be noted here: i) *pro* is licensed by agreement inflection in Null Subject Languages (NSLs, see e.g., Bennis 2006, Neeleman and Szendrői 2007, Koeneman and Zeijlstra 2014), and ii) in nonNSLs, *pro* is licensed by a functional head of a higher functional projection (see e.g., Downing 1969, Beukema and Coopmans 1989, Jensen 2003, Bennis 2006, Zanuttini 2008).[1]

There are several proposals about exactly which functional head is involved in the licensing and interpretation of *pro*: a tenseless INFL (Beukema and Coopmans 1989), an imperative-flavoured-T (Jensen 2003), an imperative C (Bennis 2006), or a Jussive T (Zanuttini 2008, Zanuttini et al. 2012). The proposal advanced here is in line with these proposals, in that it attributes a role of an interpretive import to a (higher) functional head, but it differs from them in many and various aspects, the most important of which are: i) the functional head, licensing and determining the interpretation of *pro*, is Top$^o$ in the C-domain. This head is endowed with an [Adrs] feature which yields a discourse property, ii) the (pro)nominal showing up preverbally is a topic, iii) the licensing of *pro* and its interpretation depends on coreferentiality between this topic in Spec,TopP and *pro* in Spec,*v*P, iv) *Agree* is taken as *Match,* v) our analysis employs the recent theoretic notions *Phase*, *Feature Matching* and *Feature Inherence*, (long-distance) *Agree*, etc., vi) it tackles and accounts for structures like (1d,e), and more importantly, vii) it tackles and accounts for A'-chains established in imperative structures, locally and nonlocally, a property that has never been tackled before.[2]

Discoursally, on the other hand, it has been recently realized that the (pro)nominals in (1a) and (1c) are enforced by performative/interpretive requirements, including, for instance, "the authority of the speaker over the addressee" as in (1a) (Portner 2007: 361), or the identification of the addressee as in (1c) (cf. van der Wurff 2007). This suggests that *you* and *boys* in (1a) and (1c), for instance, are required by a performative function or interpretive import, which makes explicit a discourse property (see Shormani 2017, for a comprehensive discussion). If so, these (pro)nominals cannot be considered subjects of imperatives, but rather discourse-based elements, simply because subjects do not have such properties (cf. Borer 1986, Cardinaletti 1995, Shormani 2017). We take such overt (pro)nominals to be discourse topics, specifically aboutness topics (cf. Trecci 2006, Frascarelli 2007, Shormani 2017).[3]

[1] The following abbreviations are used throughout this article: 1, 2, 3 = first, second and third person, respectively, Acc = Accusative, Adrs = Addressee, Agr = agreement, ASP = aspect, C = complementizer, EF= edge feature, EPP = extended projection principle, F = feminine, Gen = Genitive, Gend = gender, Nom = Nominative, NSLs = Null Subject Languages, Num = number, P&P = Principles and Parameters, PL = plural, S = singular, , Spec = specifier, Spcty = specificity, SVO = subject verb object, T = tense, TOP = topic, TopP = topic phrase, u = unvalued, UG = Universal Grammar, V = verb, *v* = *v* in *v*P, v = valued, VOC = vocative, VSO = verb subject object. Other abbreviations and/or acronyms used in the text are introduced in the first use.

[2] In the literature four generalizations have been confirmed in linguistic surveys on imperatives cross-linguistically: imperative clauses are universal, imperative predicates are limited to controllable processes, imperative subjects are restricted to second person and imperative subjects are optional (Alcázar & Saltarelli 2014).

[3] These authors argue that what seems to be subjects are but topics. For example, Shormani (2017) stresses that these topics are aboutness topics. They are aboutness topics in the sense that they tell us what the sentence is *about*. From a discourse perspective, we argue that the aboutness topic is the logical subject of imperatives; it is the entity that performs the speech act expressed by the imperative construction, i.e., the imperative verb plus its object. It is also the constituent that the imperative predicate is *about*. Put differently, the imperative construction is *given about/performed by* this aboutness topic (cf. Reinhart 1981, Givon 1983, Krifka 2001). The discourse link takes place via the interpretive and performative functions imperative constructions perform.

Another discourse property of imperatives can be observed in (1d); the topic cannot be iterated, specifically if the addressee is the same in both coordinate sentences, given 'Grice maxim of quantity' (see Grice 1975).[4] In (1e), on the other hand, the second topic can and must be overt in coordination. It is so because the addressee here is not the same in both coordinated imperatives; in the first coordinate, the addressee is *boys,* in the second *girls*. If the subject of the imperative is *pro*, and if the lexical NPs are topics posited in the C-domain, then, we expect two types of A'-chains, namely local and nonlocal, to be established between the topic and *pro*. Local A'-chains are established between, for instance, the topic *you* and *pro* in a local domain as in (1a), or between the topic *boys* and *pro* as in (1c). Nonlocal A'-chains are formed between the topmost topic, a null topic and *pro*(s) across sentences as in (1d). It follows that the (pro)nominal topic in each example in (1, except 1b) is the discourse antecedent of *pro*, i.e., the discourse/world entity that *pro* refers to. In the case of core imperatives (1b), we propose that a null topic (*pro*) be (re)merged in Spec,TopP, whose interpretation depends on the discourse.[5]

Yemeni Arabic is a dialect of Arabic spoken in Yemen. This language variety has several subdialects such as Sna'ani, Adeni, Ibb, Thamari, etc. In this study, I focus on Ibbi dialect, which is spoken in Ibb Governorate. Although Yemeni Arabic in general is not investigated or by linguists like many other modern varieties of Arabic, there are few studies addressing some aspects of the grammar such syntax, phonology, morphology, etc. in some dialects of Yemeni Arabic such as San'ani, Adeni, Hadhrami. However, Ibbi Arabic dialect is a very under-investigated language variety (see also Shormani 2019). Linguistically speaking, Ibbi dialect is a very rich language variety, phonologically, morphologically, syntactically, discoursally, etc. It involves linguistic "secrets" that are of much importance to the field, and that should be revealed and investigated by linguists interested in deeply probing its linguistically rich phenomena.

This article is set up as follows. Section 2 discusses the syntactic bases of imperatives, specifically imperative subjects and tense. Section 3 presents the discourse bases, underlying the formation of imperative structures and their interpretation. Section 4 briefly presents an account of the core operations of derivation in minimalism, and the notions *Feature Matching* and *Feature Inheritance*. Section 5 argues that the interpretation of *pro* is obtained in the syntax-discourse interface, building mainly on Pesetsky and Torrego's (2007: 265) assumptions that interpretation is "a requirement imposed by the interfaces between the syntax and neighboring systems", and Chomsky's (2008: 135) postulations of "the interface systems that enter into the use and interpretation of expressions". We propose that the interface between the syntax and discourse takes the form of coreferentiality between the topic(s) in Spec,TopP

---

[4] Andrew Radford (personal communication) rightly observes that *Ali* cannot be iterated in (1d) "unless you are referring to a different Ali." He points out that such a phenomenon is a "lexical economy – don't repeat words unnecessarily." He stresses that even pronouns are not allowed to iterate, providing evidence from English for the marginality/unacceptability of such imperative structures, as illustrated in (i).

(i) You sit down, and don't (? you) say a word!

Radford holds that (i) is marginal/unacceptable if 'you' is iterated, ascribing this marginality/unacceptability to "Grice's maxim of quantity (say neither more nor less than required)".

[5] The role of discourse in not saying more than needed has also been stressed in the literature. For example, Sigurðsson & Maling (2010: 77) propose what is so-called Empty Subject Condition as "reminiscent of the other empty left edge phenomena." They argue that "subjects must never be spelled out ... in those languages where infinitives (rarely) or imperatives (more commonly) otherwise allow overt subjects" as illustrated in (i) and (ii) for English and French:

(i) Take three eggs. (*You) beat __ well while someone else mixes the flour and the butter.

(ii) Prenez trois oeufs. (*Vous) deposez __ dans un bol. (*Vous)
take three eggs. you break into a bowl. you
battez __ doucement.
beat gently

and *pro*(s) in Spec,*v*P. Section 6 concludes the paper and presents some implications of the proposal advanced here.

## 2. Syntactic properties

Imperatives (both positive and negative) are considered a clause-type, and have been studied extensively cross-linguistically (see among many others, Thorne 1966, Akmajian 1984, Downing 1969, Beukema and Coopmans 1989, Zhang 1990, Merin 1991, Han 1998, Platzack and Rosengren 1998, Moon 2001, Jensen 2003, Rupp 2003, Bennis 2006, Zanuttini 1991, 2008, Radford 2009, Zanuttini et al. 2012; Shormani and Alhussen 2024). In this section, we address three important issues concerning the syntax of imperatives, namely their subjects, T(ense) and feature specifications.

### 2.1. Imperative subjects

In the literature (see e.g., the works just cited), it is assumed that the subject of imperatives can be an overt 'you' (or a lexical NP), or a null 'you' (cf. also (1)) (cf. also Radford 2009). Consider the following examples.

(2) a. You open the door!
    b. Open the door!

It was assumed that in (2a) the subject is 'you', but this assumption is not borne out, at least in Arabic, as we will see in Section 3. As for the time being, we will just address the syntactic properties of (2b). As pointed out earlier, (2b) represents the core imperatives across languages, where the subject is understood as a silent (null) pronoun. However, which null pronoun? There are two types of null pronouns in human languages, viz. PRO and *pro*. As for the subject of imperatives, there are actually two proposals in the literature: it is PRO (see e.g., Han 1998) or *pro* (see e.g., Beukema and Coopmans 1989, Jensen 2003, Rupp 2003, Bennis 2006).

In what follows, we argue that the subject of imperatives is a 2 *pro*. One piece of evidence supporting this assumption comes from the syntactic activeness *pro* exhibits. This syntactic activeness is manifested by binding/control phenomena, as shown in (3), from YA.

(3) a. đaʕ *pro* nafs-ak fii l-makaan l-munaasib!
    put.2MS self-your in the-place the-suitable
    'Put yourself in the suitable place!'

  b. ʕaawinuu *pro* baʕđ baʕđ
    help.2MPL each each
    'Help each other.'

  c. ʔint-aširuu *pro* fii l-makaan kulluh
    reflex-spread.2MPL in the-place all.it
    'Spread yourselves in the (whole) place.'

  d. rawwiħuu *pro* jazʕah la-l-bayt
    go.2MPL walking.on.foot.PL to-the-house
    'Go home walking!'

In (3), *pro* controls the anaphora in (3a), the reciprocal in (3b), the internal anaphora in (3c), and the secondary predicate in (3d). The activeness of *pro* seems to be manifested cross-linguistically, and even in non*pro*-drop languages. For example, although English and Dutch are non*pro*-drop languages, *pro*'s syntactic and semantic activeness is reflected in examples like (4) (from Bennis 2006: 108).

(4) a. Geef *pro*$_i$ elkaar$_i$ de hand!
give each other the hand
'Give each other a hand!'

b. Herinner *pro*$_i$ jullie$_i$ het gesprek van vorige week
remember yourselves the conversation of last week
'Remember last week's conversation!'

In these Dutch examples, *pro* binds the reciprocal in (4a) and the anaphora in (4b).

A word concerning the subjecthood of *pro* is in order here. Although PRO may also be assumed to be the subject of imperatives (cf. Han 1998), it cannot be a competent "rival" to *pro* in imperatives. This ensues from PRO's behavior; PRO is assumed to be in complementary distribution with a finite T (see e.g., Mohammad 2000, Roberts 2010a), in that it occurs only in nonfinite clauses. This is illustrated in (5) from English (cf. also Zanuttini 2008: 188).

(5) a. He$_i$ wants PRO$_i$ to go home.
b. *He$_i$ hopes that PRO$_i$ goes home.

In (5), PRO occurs with the infinitive *go*, because T in infinitives is nonfinite. However, it cannot occur in finite clauses as the ungrammaticality of (5b) shows. As we will see in the next section, T in imperatives is finite, which gives us a room to postulate that, on the one hand, PRO cannot be assumed to be the null subject of imperatives, and on the other hand, *pro* may be the only available alternative (see also Beukema and Coopmans 1989, Bennis 2006).[6]

Chomsky (1995: 106-109) distinguishes *pro* from PRO, arguing that *pro* "typically occurs as the subject of a finite clause" but PRO cannot occur in such a position. He attributes this to the fact that *pro* has Case, though he has also assumed that "PRO, like other arguments, has Case, but a Case different from the familiar ones: nominative, accusative, and so on." However, PRO is again distinguished from *pro,* in that while the latter can move from a Case-marked position (to another Case-marked position), "PRO is permitted to move from a non-Case position to a position where its Case can be assigned or checked, and is not permitted to move from a Case position." Chomsky also argues that PRO is "a "minimal" NP argument, lacking independent phonetic, referential, or other properties." In addition, Kratzer (2009: 189, fn.2) adds another difference between both constituents, holding that PRO differs from *pro,* in that the former can be a "minimal pronoun", but the latter "does not have to be. Like its overt counterparts, *pro* can be born with all its features in place, in which case it is referential" (see also Landau 2013, for very recent conceptions on PRO distribution and differences). Given this, and as far as we can tell, PRO may not exist in Arabic. This stems from the fact that clauses in Arabic are always finite, even control clauses. Consider (6) from Standard Arabic.

[6] For example, based on *Principles and Parameters* (P&P) assumptions, Beukema and Coopmans (1989: 420-422) discuss four alternatives of null categories that could be assumed as subjects of imperatives, namely trace of NP-movement (NP-t), trace of *Wh*-movement, *(Wh-t),* PRO and *pro*. They exclude the NP-trace on the bases that it requires an identification by an antecedent in a c-commanding argument position, which is not available. Likewise, they exclude the *Wh-t* as there is clearly no PF/i-antecedent in imperatives. They also exclude PRO based on the ground that it cannot occupy a Case-marked position, i.e., Spec,INFL, because it is governed by INFL. They conclude that *pro* is the only alternative that can be taken as the subject of imperatives.

(6) a. ʔaraada ʔan yaktuba *PRO/*pro* d-dars-a
wanted.PAST.he C write.PRES.he the-lesson-ACC
'He wanted to write the lesson.'

b. ʔan qad jaaʔa *PRO/*pro* min r-riħlat-i
C may came-PAST from the-trip-GEN
'He may have come from the trip.'

As can be seen in (6), the verb in Arabic is always finite; though used in control structures. In (6a), for example, the verb *yaktuba* is in the present tense and in (6b), the verb *jaaʔa* is in the past tense. Compared to (5), it is clear that not only are Arabic verbs inflected for tense, but also for φ-features (see also e.g., Shormani 2015, for discussions).

### 2.1.1. *pro's* licensing in imperatives

It has long been recognized that human language does not allow saying more than necessary, based on information exchange (see e.g., Grice 1975, Chomsky 1981, Kiss 1995, Erteschik-Shir 2007). If 'you' is the subject of imperatives, as assumed cross-linguistically (see e.g., Ackema et al. 2006, van der Wurff 2007), it follows that "dropping" it will not affect information exchange, given 'Grice's Maxim of quantity'. But dropping 'you' does not mean that the imperative subject is eliminated from the syntax, but rather suggests that it is covert, i.e., 'have a null spellout', which conforms to the Avoid Pronoun Principle (APP) (see Chomsky 1981).

The *pro*-drop phenomenon has received much research in syntactic theory (see e.g., Jaeggli and Safir 1989, Rizzi 1982, 1986, Huang 1984, 1989, N. Hasegawa 1985, Neeleman and Szendrői 2007, Holmberg 2005, Ackema et al. 2006, Biberauer et al. 2010, Koeneman and Zeijlstra 2014). In terms of *pro*-drop allowance or disallowance in declarative clauses, human languages can be divided into three typologically distinct categories, namely i) *pro*-drop languages like Arabic, Italian, Finnish, Hebrew, etc., ii) *pro*/topic-drop languages like Chinese, Japanese, Korean, etc. and iii) non*pro*-drop languages like English, French, Dutch, etc. As for imperatives, however, all these language types seem to allow *pro*-drop phenomenon. Consider the following examples from the three categories each.

(7) a. ʔiktub/-uu *pro* d-dars
write.2MS/-2MPL the-lesson (Arabic)
'Write the lesson.'

b. Cemsim-ul sa-la!
lunch-ACC buy-IMP
'Buy lunch!' (Korean, Zanuttini et al. 2012: 1240)

c. (Tu) écris!
(You.SING) write
Write! (French, cf. Rowlett 2014)

d. '(You) write!' (English)

As can be observed, imperative force renders all human languages to have *pro*-drop property. It also seems that 'rich agreement' morphology has nothing to do with this property in imperative constructions. This gives us enough room to postulate that discourse does have a role to play in *pro*-drop phenomenon in imperatives. Huang (1984, 1989) proposes that *pro* is always controlled by a (null) topic. Following Huang, Moon (2001) proposes that discourse actually underlies the choice between *you* and *pro*. She argues that imperatives are subjectless in nature, suggesting that the overt (pro)nominal imperative subject is actually a topic-prominent constituent. Neeleman and Szendröi (2007); moreover, propose that dropped pronouns are regular pronouns, but they receive phonologically null Spell-Outs. However, the question is: what is exactly the discourse force that allows imperatives to drop their subject? (I return to this issue in Section 3).

## 2.2. Tense and T's status in imperatives

It was assumed that imperatives lack tense, hence TP projection altogether (see e.g., Beukema and Coopmans 1989, Zanuttini 1991, Platzack and Rosengren 1998). Imperatives were viewed as not exhibiting tense contrast- they have only one verbal form (van der Wurff 2007). This is, however, not borne out as we will see in this section. In particular, we show that imperatives have tense, hence T. We also argue that T in imperatives has φ-features similar (but not identical) to those of T of other clause types.

Imperatives may be assumed to have a tense of some sort. This tense might be present or future, but not past, as illustrated in (8), from YA.[7]

(8)

a. taʕaal baʕd saaʕah ʕa tjah?
   come after hour will you.come
'Come after an hour, won't you?'

b. *taʕaal baʕd saaʕah qa jiʔk?
   come after hour ASP came.you

The imperative structure in (8a) consists of an imperative and a question tag. In YA, *ʕa* is a future particle, but *qa* is a perfect/past one. (8b) is ungrammatical and this ungrammaticality obviously indicates that past tense is not possible in imperatives. However, the very idea that (8b) is impossible suggests that there is some sort of tense feature associated with T in imperatives that prevents such a structure (cf. Shormani and Qarabesh 2018).

This analysis is in line with Jensen's (2003) proposal who proposes that imperatives have a tense feature, but different from that of declaratives. She ascribes this tense feature "to the presence of an imperative-flavoured-T° that competes with prototypical-declarative-T° for this functional position" (p. 158) (see also Rupp 1999).

[7] This is also supported by Arabic traditional grammarians (see e.g., Sibawayhi 1938: v.1, Wright 1898) who consider that the tense of imperatives is present or future, but not past. They argue that the actual "happening" of the speech act performed by an imperative verb is being carried out after or at the time of speaking. They actually base their argument on some temporal adverbial modification as in *do it now/tomorrow*, but not *do it yesterday*, for example.

There is also a strong piece of evidence in support of the idea that imperatives have tense, ensuing from languages like Latin as in (9) (from Baldi 1999: 404, cited after van der Wurff 2007: 43).

(9)
a. ī!
go.IMP.SING.PRES
'(You.SING ) go!

b. īto!
go.IMP.SING.FUT
'(You.SING ) go!

Latin seems to mark tense in imperatives, but again, the tense is either present as in (9a) or future as in (9b). Based on these facts, we propose that T has a present/future tense.

**2.2.1. T's feature specifications**

If imperatives have tense (hence T), then, a question arises as to whether this T has the features akin to T of other clause-types, say, declaratives. In declaratives, T has φ-features, Case and EPP features. As for φ-features, YA provides empirical evidence that T in imperatives has these features. Consider the following examples.

(10)
a. ʔiktub *pro*!
write.2MS

b. ʔiktub-i *pro*!
write-2FS

c. ʔiktub-uu *pro*!
write-2MPL

d. ʔiktub-ayn *pro*!
write-2FPL

As the gloss shows, T in YA imperatives seems to be φ-*complete* in the sense of Chomsky (2000, 2001). However, one crucial difference between T of imperatives and that of other clause-types is that T in imperatives seems to have a 2 person feature cross-linguistically.[8]

As for Case, we assume that T enters the derivation with a Nom Case feature. Under the proposed analysis, it is expected that *pro*'s unvalued Case feature is valued by the valued Case feature of T. We also assume that T has an EPP feature. If we take Arabic, be it standard or YA, as a SVO language, then under the analysis pursued here, *pro* needs to merge in Spec,*v*P, and then raises to Spec,TP in order to value T's EPP feature. If, however, we take it as a VSO language (which we adopt in this article, see e.g., Shormani 2015), then while *pro* merges and stays in situ, i.e., Spec,*v*P, the EPP of T could be assumed to be valued via V-to-T movement

[8] Examples in (10) and throughout lend us support that imperative verbs are finite as indicated by the fact that they inflect for all φ-features. This also leads to the assumption that imperative verbs have an absolute Tense interpretation (cf. Bianchi 2003).

(cf. Alexiadou and Anagnostopoulou 1998: 494, Koeneman and Zeijlstra 2014 كٍاخقةشه 2025).[9]

Another difference that can be noted here is that imperative T’s tense feature is unchangeable unlike the declarative/interrogative T. That is, it is always present/future (see also Zhang 1990). It turns out, then, that the imperative T has φ–features, tense and Case features similar (but not identical) to those of T, say, in declaratives. However, the question is: where do these features come from? We address this question in Section 4.

## 3. Discourse properties

In this section, we discuss two discourse-related properties of imperatives, namely: i) imperatives have performative functions, and ii) their subject is controlled/bound by an NP in the C-domain.[10]

Imperatives have been used, in several and different ways, to express different and various functions. The fact that imperatives are related to discourse, perhaps more than any other clause-type, comes from ‘their very dialogic nature’. Consider the following examples from YA.

(11)

a. tafađđal ʔijlis!
please sit.2MS
‘Please, have a seat!’

b. ʔintabih!
be.careful.2MS
‘Be careful!’

c. ʔitkallam bi-ʔadab!
speak.2MS with-politeness
‘Speak politely!’

d. yaa ʔallah, ʔirħam-naa!
VOC God, have.2MS.mercy.on-us
‘O God, have mercy on us!’

e. ʔiftaħ l-baab
open.2MS the-door
‘Open the door!’

f. ðaakiruu bi-jidd!
study.2MPL with-hardiness
‘Study hard!’

g. quum!
stand.2MS
‘Stand up!’

[9] The idea that T has an EPP feature is minimalist in nature, simply because it makes dealing with imperatives like dealing with other clause-type structures, which gives rise to nonconstruction-specific postulations (cf. Jensen 2003). Furthermore, this assumption makes our proposal have a cross-linguistic appeal, in that it can be applied to VSO languages like Arabic, Irish, Hebrew, etc., and SVO languages like English, French, Italian, etc.

[10] There are also other approaches to the interpretation of imperatives like the modal approach and the semantic approach, for more on this (see e.g., Han 1998, Kaufmann 2012). Another approach to imperatives that can be noted here is *To-do-list* proposed by Portner (2007). However, these approaches differ in scope and content from the one proposed here.

From a discourse/pragmatics point of view, imperatives as in (11) are deemed to perform various speech acts like command, order, advice, request, etc. (see e.g., Zhang 1990, Han 1998, 2001, Aloni 2007, Aikhenvald 2010). For example, Zhang (1990: 11) identifies several ways in which imperatives are used. Syntactically, an imperative can mean "a class that is parallel to declarative and interrogative". Pragmatically/discoursally, it can mean "the pragmatic notion of directive, including commanding, ordering, advising, requesting, suggesting, that is parallel to notions such as assertives, expressives and so on" (cf. also Bianchi and Frascarelli 2010). Along these lines, Huddleston (2002: 929) argues that imperative structures are typically used as directives, expressing 'a proposition representing a potential situation: realizing or actualizing that situation constitutes compliance with the directive.' Aloni (2007:70) identifies a crucial difference between imperatives, on the one hand, and declaratives and interrogatives, on the other. Aloni argues that declaratives have truth conditions, interrogatives have answerhood conditions, while imperatives have compliance conditions. She also argues that it is difficult for someone "to understand the meaning of an imperative unless she recognizes what has to be true for the command (or request, advice, etc.) issued by utterance of $!\phi$ to be complied with."[11]

As noted above, the performative functions imperatives have are implemented in performing speech acts like respect, threat, command, request, order, etc.[12] In (11a-g) the speech acts expressed are respect, warning, threat, prayer, request, advice and order, respectively. The fact that discourse plays an important role in the interpretation of an imperative sentence is clearly manifested in (11). For example, the situation in which (11a) is said might be that of a student and a teacher, whereby the former is the speaker and the latter the addressee. It may also be that of an employer and an employee, a son/daughter and a father, a clerk and a manager, and so forth, depending, of course, on who takes the conversational role in such situations.

Examples like (11b) may be said in a teacher-student care situation. It may also be said in a threat situation, where a teacher threats his or her male student to study hard so that he may not fail in the exam. It may also express a type of advice. In fact, each sentence in (11) may express/perform several speech acts, still, depending on the discourse, however. The idea that discourse plays a crucial role in the interpretation of imperatives has also been realized in the literature (see e.g., Aikhenvald 2010, for a cross-linguistic data, including English, and Vlemings 2003, for French).

One piece of evidence of the discourse involvement in interpreting imperatives comes from vocative structures like (11d), repeated here as (12a) (cf. Shormani and Qarabesh 2018, Shormani 2020).

(12)
a. yaa ʔallah, ʔirħam-naa!
VOC God, have.2MS.mercy.on-us
'O God, have mercy on us!'

b. ʔirħam-naa!
have.2MS.mercy.on-us
'Have mercy on us!'

[11] The symbol $!\phi$ is mentioned in Aloni (2007: 86) "where '!' is an operator over the set of propositional alternatives introduced in its scope."

[12] Han (2001: 290, emphasis in the original) observes that "the term *imperative* has often been used to refer to a sentence's function rather than its form".

If, however, (12a) is said without the vocative phrase *yaa ʔallah,* as in (12b), the sentence will be ambiguous, and the speech act expressed by the imperative would not mean *prayer*, but rather *begging/appeal*. If we take into account the current views on vocatives that they are part of the C-domain (see e.g., Hill 2007, 2013, 2014, Stavrou 2013, Corver 2008, Sonnenhauser and Noel Aziz Hanna 2013, Haegeman 2014; Alcázar and Saltarelli 2014, Shormani and Qarabesh 2018, Shormani 2020, 2024a & b), and given that the C-domain represents the informational coda, i.e., discourse (cf. Vallduví 1992, Lambrecht 1994, Rizzi 1997, *et seq*, Cinque 2006, Cinque and Rizzi 2010, Shormani 2017), it follows that discourse plays a crucial role in the interpretation of imperatives.[13] Given our assumption that *pro* is the (thematic) subject of imperatives, and that this *pro* is bound by an NP in a higher position, then, *ʔallah* in (12a) could be assumed to be the antecedent of *pro*. Note that when we remove the vocative particle *yaa* from (12a), and write the sentence as (13) below, then, *ʔallah* in (13) seems to be a topic, i.e., a left dislocated element, more than a vocative. The idea that vocatives are *extremely* similar to topics has been advocated in the literature (see e.g., Lambrecht 1996, Portner 2004, see also Alcázar and Saltarelli 2014, Shormani and Qarabesh 2018, Shormani 2020).

(13) ʔallah, ʔirħam-naa!
God have.2MS.mercy.on-us
'God have mercy on us!'

Bearing all this in mind, we are now in a position to prove that the (pro)nominal constituent showing up preverbally in imperative constructions is a topic. If (11e) above, repeated here as (14b), is viewed as the core structure of imperatives cross-linguistically, then, it is tempting to postulate that *ʔantah* in (14a) is not the subject, but rather a left dislocated element, i.e., a topic (cf. Koopman 2007).[14]

(14) a. ʔantah ʔiftaħ *pro* l-baab
you open.2MS the-door
'You, open the door!'

b. ʔiftaħ *pro* l-baab
open.2MS the-door
'Open the door!'

The Arabic imperatives in (14a & b) provide crucial evidence in support of the assumption that *ʔantah* in (14a) is a topic rather than a (thematic) subject. Strong evidence in support of our assumption is provided in (15), from YA, where *ʔantah* occurs postverbally.

ʔiftaħ ʔantah *pro* l-baab

[13] The idea that the overt subject of imperative is a vocativized nominal dates back to the work of Thorne (1966) Thorne (1966:74f, emphasis in the original) argues that "the SURFACE STRUCTURE subject of an imperative sentence must be vocative." He also stresses that the "vocative forms are obligatory in the subject of all imperative sentences." One problem in Thorne's analysis, however, is that he seems to consider all imperatives, though in English, to always have overt subjects, which is not unproblematic (see e.g., Zhang 1990, Moon 2001, for a criticism).

[14] Koopman (2007) discusses topics in Dutch and German imperatives; she does not address topics from a subjecthood perspective, but rather tackles the right and left periphery phenomenon of objects, right or left dislocated. In particular, she concerns herself with objects, and when they can or cannot be topicalized in imperatives.

(15)
open.2MS you the-door

'You, open the door!'

Further evidence in support of our postulation that *ʔantah* in (14a) is a topic is to substitute *ʔantah* with a lexical NP, as illustrated in (16).

(16) ʕali ʔiftaħ *pro* l-baab
Ali open.2MS the-door
'Ali, open the door!'

The lexical NP *ʕali* in (16) functions as a topic (cf. Lambrecht 1994, 1996). This amounts to the fact that it is 'posited' in the C-domain, or in Spec,TopP in a Rizzian-1997 clause architecture.

One common property of (14a) and (16) is that *ʔantah* and *ʕali* are both understood as the addressees (cf. e.g., Lambrecht 1996, Portner 2004, Zanuttini 2008). This explicitly suggests that *ʔantah* and *ʕali* are not subjects, but rather topics (cf. Reinhart 1981, Givon 1983, Krifka 2001, Erteschik-Shir 2007); and therefore, belong to the informational coda in Vallduví's (1992) sense. If informational coda is equated with discourse (see e.g., Rizzi 1997, *et seq*; Krifka 2001, Erteschik-Shir 2007; Shormani 2017), then, *ʔantah* and *ʕali* belong to discourse.[15]

It turns out that discourse plays a crucial role in the presence of a topic in imperatives. That is to say, if the presence of a topic in imperatives is enforced by a performative/interpretive requirement, say, performing an interpretive import, or initiating a speech act, it is reasonable to postulate that the selection of this topic is determined by the speech act itself. Along these lines, Krifka (2001: 25) argues that the speech act itself plays a role in selecting topics, "an initiating speech act that requires a subsequent speech act like an assertion, question, command, or curse about the entity that was selected". Krifka's argument is at the heart of our proposal. The topic in our story is assumed to initiate a speech act, which requires "a subsequent speech act", in Krifka's sense, to be carried out by the imperative structure.

### 3.1. Aboutness Topics

However, the question is: which topic is intended in our system? Topics have cross-linguistically been classified into several types such as *aboutness topic, contrastive topic, familiarity topic, givenness topic,* etc. (for details, see e.g., Frascarelli and Hinterhölzl 2007, Bianchi and Frascarelli 2010, Shormani 2017, 2024a & b; Shormani and Alhussen 2024). The topic proposed in this study is *aboutness topic* that counts as a logical subject, and not any other topic, or topicalized constituent for that matter.

[15] Topics have cross-linguistically been classified into two types, viz. sentence topics and discourse topics. For example, Reinhart (1981: 54) argues that while the former "must correspond to an expression in the sentence, discourse topics are topics of larger units and can be more abstract." To her, subject topics are not very much different from object topics in "sentence-discourse" contrast. Bayer (1980: 7), on the other hand, sees that subjects are more likely to be discourse topics than objects are. Bayer argues for discourse topics and takes the assumption "old vs. new information" as a criterion to distinguish subject topics from object ones. To him, the former express "old information" and "coincide more often with the intuitively felt topics than "new information" and so on" (cf. also Krifka 2001, Bianchi & Frascarelli 2010, for similar conceptions). Bayer's and Krifka's position that the subject is a topic is, in fact, at the heart of the proposal pursued here, in that while the topic will be the logical subject, the thematic subject of imperatives is *pro* (see also Shormani 2017, Shormani 2017, 2024a & b; Shormani and Alhussen 2024).

(17)

a. | ʕali | ʔiftaħ | l-baab | li-ʕaliya! |
|---|---|---|---|
| Ali | open.2MS | the-door | for-Alia |

‘Ali, open the door for Alia!’

b. | li-ʕaliya | ʕali | ʔiftaħ | l-baab |
|---|---|---|---|
| for-Alia | Ali | open.2MS | the-door |

‘Ali, open the door for Alia!’

c. | l-baab | ʕali | ʔiftaħ | li-ʕaliya |
|---|---|---|---|
| the-door | Ali | open.2MS | for-Alia |

‘Ali, open the door for Alia!’

The topic we are concerned with in this study is *ʕali*, not *li-ʕaliya*, *ʕaliya* or *l-baab*. It is an aboutness topic, having a contrastive nature in the sense of Bianchi and Frascarelli (2010). It is an aboutness topic with a focus in the sense of Portner and Yabushita (1998) and Krifka (2001, 2008). It is an aboutness topic that identifies/picks up a noun phrase as the addressee in the sense of Shormani and Qarabesh (2018). Such topics “represent a combination of topic and focus… They consist of an aboutness topic that contains a focus, which is doing what focus always does, namely indicating an alternative. In this case, it indicates an alternative aboutness topic” (Krifka 2008: 267). They can also be omitted from the (narrow) syntax due to the fact that the information strategy is already there, i.e., known to the interlocutors in a conversation (cf. Büring 2003). The possibility of “omitting” aboutness topics from imperative constructions strongly supports our proposal concerning “subjectless” or core imperatives (I return to this point in section 5).

Thus, in our examples in (17), for instance, the aboutness topic identifies/picks up *ʕali* from a set of topics, namely *li-ʕaliya*, *ʕaliya* and *l-baab* in (17a-c) and contrasts it with these topics in the sentence; it is *ʕali* and nobody else who will open the door for *ʕaliya*. It is neither (li-)*ʕaliya* in (17b), nor *l-baab* in (17c), or any other topicalized constituent for that matter (cf. Erteschik-Shir 2007). It is a base-generated element in the sense of Shormani (2017), and has its own features. These features are by and large a result of being a discourse constituent. The aboutness topic in our story is one which has scope over the whole proposition (cf. Frascarelli 2007, Shormani 2017). In this sense in (17a), for instance, the aboutness topic *ʕali* has scope over the whole proposition of the imperative sentence.

One and the most distinguishing feature of aboutness topics is that they are not recursive in the sense that they are not iterated (cf. Frascarelli and Hinterhölzl 2007, Bianchi and Frascarelli 2010, Shormani 2017). I adopt here these authors’ ideas and apply them in our analysis. In fact, this “nonrecursiveness” is a substantial feature which characterizes aboutness topics as logical subjects, because subjects are not iterated (see e.g., Borer 1986, Shormani 2015), thus strongly supporting our proposal that the logical subject of imperatives is an aboutness topic.

(18)

| *ʕali, | ʔaħmad, | ʔiftaħ | l-baab | li-ʕaliya! |
|---|---|---|---|---|
| Ali, | Ahmed | open.2MS | the-door | for-Alia |

In (17a), *ʕali* co-occurs with other topics, namely *li-ʕaliya* in ((17b), and *l-baab* in (17c). However, comparing (17a) to (18), it seems that the only reason for the ungrammaticality of (18) is the co-occurrence, i.e. the recursiveness, of the aboutness topic *ʕali* with another aboutness topic, viz. *ʔaħmad*.[16] I will further discuss the notion of (non)recursiveness in the clausal imperative projection in the following section.

**3.2. Informational structure**

Cartography-based approaches to projection assume that features and information coding factors have structural representations and project their own projections. This informational structure is invariably the C-domain (see e.g., Vallduví 1992, Lambrecht 1994, Benincà 2001, Rizzi 1997, 2004, 2006, Erteschik-Shir 2007, Shormani and Qarabesh 2018). The C-domain has been seen as representing root-clause properties like interrogativeness, declarativeness, imperativeness, and as encoding discourse features like topichood, force, focus, etc. In the words of Rizzi (1997:283), the C-domain is "the interface between a propositional content (expressed by IP) and the superordinate structure (a higher clause or, possibly, the articulation of discourse, if we consider a root clause)." Rizzi (1997) proposes that C-domain is further split into three functional categories, namely ForcP, TopP, and FocP (in addition to FinP). Thus, we propose (19) as the structure of imperatives.

(19) TopP … (ForcP) … (FocP) … TP

In (19), TopP is the topmost projection, followed by ForcP, which is in turn followed by FocP, and TP. The proposal in (19) gains support from imperative structures in YA as illustrated in (20). It also gains support from English as shown by the translation of these examples.[17]

(20) a. ʕali, ʔayna, ʔantah? taʕaal!
Ali, where you come
'Ali, where are you? Come!'

b. ʕali, ʔinna ʔallah maʕak!
Ali C God with.you
'Ali, God be with you!'

(21) a. ʕali, muh l-kitaab taʕaal!
Ali, is the-book come
'Ali, is the book with you?'

b. *ʕali, muh ʔinna al-kitaab maʕak, taʕaal!
Ali is C the-book with.you come

(22) a. *ʔayna ʔantah? ʕali, taʕaal!
where you Ali, come

b. *muh/ʔinna *ʕali, l-kitaab maʕak, taʕaal!
Is/C Ali the-book with.you come

[16] Note that the topics in our analysis cannot be "pure" contrastive topics (CT) due to the fact that contrastive topics can be recursive (see e.g., Büring 2003). Büring found examples of CT + CT constructions from German and some other languages.

[17] A reviewer asked whether *ʕali* in (20a &b) can be a vocative rather than a topic. As the article suggests, there is some sort of interchangeability between topics and vocatives. However, I consider the occurrence of *ʕali* in this context more likely to be a topic, rather than a vocative. One simple reason could be the absence of the vocative article (see also Shormani 2017, 2020).

These examples show that TopP, but not ForcP, is the topmost projection. Given that wh-words (re)merge in Spec,ForcP, (20a) supports our postulation.[18] This is farther supported by (20b). In (20b), the complementizer, i.e., ʔ*inna* is merged in Forc°.[19] The fact that ʔ*inna* is merged in Forc° gains support from the declarative nature of the embedded clause in these imperative structures (cf. Ross 1970, Chomsky 1995).[20] One strong piece of evidence supporting this assumption comes from the complementary distribution between ʔ*inna* and *yes-no* question particles like *muh* in (21a), thus rendering (21b) ungrammatical. As a question particle, *muh* is merged in Forc° (see also (Benincà 2001: 62), for Italian TopP and ForcP positioning in declaratives, and (Shormani 2017), for such positioning in Standard Arabic). Furthermore, the ungrammaticality of (22a,b) also adds strong support to the reliability of the proposal in (19), at least for imperative structures, and indicates that TopP must precede ForcP. Finally, (22b) shows that the only possible position ʔ*inna* or *muh* can occupy is Forc°, which also supports (19).[21] Thus, we are now ready to address the feature specifications of the head Top° in our system, and we tackle these in the following section.

**3.1.2. Top's specifications**

The TopP, headed by Top°, can be taken as a phase in the C-domain. If so, Top° must exhibit the characteristics of phase heads in general, i.e., it must have the feature-composition of C, viz. φ-features and Tense (and Case) (cf. Chomsky 2001, *et seq*). As for the former, Standard Arabic provides independent empirical evidence from relativized constructions that C has φ-features as (23) shows, where C (i.e., the relative pronoun) agrees with the DP it introduces in all φ-features (see also Shormani 2017).[22]

| | | | | |
|---|---|---|---|---|
| a. | r-rajulu | llaði | jaaʔ-a | *pro*. |
| | the-man.3MS | who.3MS | came-3MS | |
| | 'The man who came' | | | |
| b. | r-rijaalu | llað-iina | jaaʔ-uu | *pro*. |
| | the-man.3MPL | who-3MPL | came-3MPL | |
| | 'The men who came' | | | |
| c | al-bintu | llati | jaaʔ-at | *pro*. |
| | the-girl.3FS | who.3FS | came-3FS | |

[18] Note, however, that the proposal in (19) is different from Rizzi (1997), in that in our proposal topics are not recursive. (19) is in fact in line with Benincà (2001, *et seq*). Note also that (19) and its conceptions are supported by the fact that the topic is the logical subject of imperatives, which is not iterated (recursive), and other types of topics such as F-topic and G-topic are not part of our system. Unlike Rizzi, Benincá proposes that topic precedes focus and "there is not an optional position for Topic below Focus", and this has also been strengthened in her paper Benincá (2001) and subsequent work like Benincá (2006). These ideas have also been approved by Rizzi (2004), specifically referring to Benincá and Poletto's (2004) hypothesis that "...topic strictly precedes left-peripheral focus..." (Rizzi 2004: 9).

[19] Along these lines, Kiss (1995: 12) points out that "[t]opics typically precede WH-phrases". She also states that "topics in embedded clauses tend to follow the complementizer."

[20] Note that Arabic has two complementizers of the sort ʔ*inna:* ʔ*inna* and ʔ*a*nna. The former can be taken as a root clause marker while the latter of embedded clause. The latter in fact comes always in embedded clauses, specifically of the type 'say-clause'.

[21] A reviewer asked: "Is the sentence [in (21b)] grammatical without *muh*? If not, then *muh* is not necessarily in the same position as ʔ*inna* (i.e., C)." The answer to this question is actually, ***yes***. The sentence is grammatical without *muh* as (i) indicates:

(i)

| | | | | |
|---|---|---|---|---|
| ʕali, | ʔinna | al-kitaab | maʕak, | taʕaal! |
| Ali | C | the-book | with.you | come |

'Ali, indeed, the book is with you, come!'

[22] See also Rouveret (2008: 190, fn.10), for data from Welsh showing that C is φ-*complete*.

(23) 'The girl who came'

d. al-banaatu llaati jiʔ-na *Pro.*
the-girl.3FPL who.3FPL came-3FPL
'The girls who came'

In (23a-d), C (*llaði, llað-iina, llati* and *llaati,* respectively) agrees with the constituent it introduces, namely *r-rajul*, *r-rijaalu*, *al-bintu* and *al-banaatu*, respectively, in all φ-features (cf. Shormani 2017). More importantly, these examples show also that C agrees with the verb, again, in all φ-features (cf. Kayne 1983, 1994, Borsley 1997, Shormani 2017).

Although tense does not show on C in Arabic, there are languages in which C shows tense feature (by means of inflection). For example, Adger (2007: 34) argues that C in Irish exhibits a past and non-past tense contrast, as illustrated in (24).

(24) a. Deir sé go dtógfaidh sé an peann.
say.PRS he that take.FUT he the pen
'He says that he will take the pen.'

b. Deir sé gur thóg sé an peann.
say.PRS he that.PST take.PST he the pen
'He says that he took the pen.'

As can be observed, C shows tense contrast; it is *go* in (24a), but *gur* in (24b). The former is present and the latter is past. Given this, it is possible to assume that T (even in imperatives) inherits C's tense feature.

Recall that we concluded in connection with the examples in (14-16) that 'you' and 'Ali' represent the addressees in these imperative constructions. And given that this addressee is merged in Spec,TopP, it follows that the head Top° is endowed with an [Adrs] feature. It also follows that (25) holds true of Top° (cf. Rizzi 2006, Frascarelli 2007, Shormani 2017).

(25) Top° is a criterial position in C-domain, and endowed with an [Adrs] feature which yields a discourse property, and links the topic with *pro* in T-domain

In terms of (25), the feature [Adrs] the head Top° is endowed with cartographically constitutes an 'information structure primitive' in the left periphery, solely needed as an information/discourse requirement. If we take the [Adrs] feature as an interpretative import, it turns out to be an *Edge Feature* (cf. Chomsky 2005, 2008: 139), which is valued via (re)merging a topic in Spec,TopP.[23] We take the [Adrs] feature to entail initiating a speech act (cf. Corver 2008: 89), which is performed by an imperative verb/structure. It is also reasonable to postulate that the feature [Adrs] correlates the discourse with syntax, i.e., the informational coda and the propositional structure, respectively.

[23] Chomsky (2005, 2008) proposes that based on LF interpretation purposes Lexical Items (LIs) enter the derivation endowed with an edge feature. This feature enables them to enter the computation. In the words of Chomsky (2008:139), "[a] property of an LI is called a feature, so an LI has a feature that permits it to be merged. Call this the edge-feature (EF) of the LI." In our system, the fact that [Adrs] feature counts as an EF is motivated by LF interpretation purposes, and stems from the behavior of the topics. The topic is assumed to be the addressee, and the logical subject of imperatives, which greatly contributes to the interpretation of imperative constructions. If the head Top° has an [Adrs] feature, then it will have two probes, namely *Agree Feature* and EF (cf. Chomsky 2008: 148). The former concerns φ-features, in that Top° probes for valuing its unvalued φ-features via *Agree* with *pro* in Spec,*v*P. The latter, however, is satisfied by (re)merging a topic in Spec,TopP.

There are two substantial properties characterizing the (aboutness) topics intended here: i) even if they are 3 NPs, they bind 2, but not 1 nor 3 *pro* in imperatives (cf. Frascarelli 2007, Shormani 2017), and ii) they must be specific in imperatives (in contrast, say, to declaratives), in that they are the addressed entities.

(26)
a. ʕali ʕarrif nafas-ak
Ali introduce self-your
'Ali introduce yourself!'

b. *ʕali ʕarrif nafas-i
Ali introduce self-my

c. *ʕali ʕarrif nafas-uh
Ali introduce self-his

d. ʔawlaad ʔimšuu, banaat ʔibquu
boys leave, girls stay
'Boys leave, girls stay!'

Although the aboutness topic *ʕali* is a 3 person NP, it binds a 2 person (anaphora) NP, namely *nafas-ak*. It cannot bind 1 or 3 NPs as evidenced from the ungrammaticality of (26b,c). These 3 NPs turn out to be 2 NPs; what gives them this characteristic, we believe, is being coreferentially linked with a 2 *pro*. Furthermore, the NPs *ʔawlaad* 'boys' and *banaat* 'girls' in (26d) are indefinite, but they each seem to be specific in imperatives. Syntactically, these NPs are made specific by being coreferentially linked with *pro*, which is a 2 person *pro* (I discuss these properties in details in section 5). Discoursally, these NPs are specific due to being part of the discourse, i.e., they each denote/refer to a specific entity in the world. For instance, the NP *ʔawlaad* picks up a group of individuals involved in the discourse of the imperative sentence (26d). Given this, the specificity of these topic-NPs seems to be a *discourse* property which individualizes an indefinite DP, making it refer to a (specific) referent (see also Uriagereka 1995, Lyons 1999).[24] This individualization makes *ʔawlaad* 'boys' and *banaat* 'girls' in (26d) refer to a specific referent 'uniquely determined for the speaker and the addressee' (cf. Lyons 1999: 59f, see also Mohammad 2000: 111ff; Giorgi 2010). This specificity could be a feature linked to [Adrs], which is, in turn, a property of the informational coda/structure (cf. Erteschik-Shir 2007, Shormani 2017). It follows, then, that the head Top° is endowed with a specificity [Spcty] feature as well.[25] We conclude that the head Top° is endowed with discourse-based features, namely [Adrs] and [Spcty], and syntax-based features, namely φ-, Case and tense features.

One more crucial point we would like to address here is that we have taken the feature [Adrs] as an EF, because it is discoursally more prominent than the [2Pers] feature. In the literature, those who assume a higher head to license *pro* take the [2Pers] feature as more prominent than any other feature (see e.g., Bennis 2006, Zanuttini 2008, Zanuttini et al. 2012). Although it is possible to assume the [2Pers] feature to be an EF of Top°, the [Adrs] feature seems to be more logical than [2Pers]. Theoretically, under the proposed analysis, the [2Pers] feature cannot

[24] Recent studies show that topics are not *always* definite. Under discourse, topics can be indefinite (see e.g., (Erteschik-Shir 2007), for Italian, and (Shormani 2017) for Arabic)

[25] The same line of reasoning can also be extended to the 2 person [2Pers] feature that Top° is endowed with. The assumption that Top° is endowed with a 2 person feature is basically supported by the occurrence of pronouns like 'you' as topics in imperative structures like (14-16) above. But since the 2 person feature is part of Top°'s φ-composition, we consider it a syntactic feature, rather than a discourse one.

*simultaneously* be assumed as a φ-feature and an EF one. That is to say, since the [2Pers] is part of the φ-*composition* of the head Top°, it is theoretically untenable to take it as an edge feature as well.

## 4. Core operations, *Feature Match* and *Feature Inheritance*

In our system, the computation procedure is taken to arrange and rearrange "items taken from the lexicon according to their properties with a view to meeting the requirements of Full Interpretation" (Boeckx 2003: 2). We take these "arrange and rearrange" as *Merge* and *Move*, respectively. As for *Merge, we* adopt Chomsky's postulation that "[f]or an LI to be able to enter into a computation, merging with some SO (and automatically satisfying SMT), it must have some property permitting this operation" Chomsky (2008: 139). This property, Chomsky argues, is a *feature*, specifically an *edge feature*. We will also take *Move* as *Copy* (cf. Fox 2002). Within that space lies another core operation, i.e., *Agree,* which 'regulates' the interaction between a probe P and a goal G. This interaction may take the following three mechanisms (cf. Chomsky 2000: 122, Boeckx 2003: 2f, Shormani 2017: 151).

(27) a. Features trigger *Match* (e.g., there is a valued and interpretable [Adrs] feature on the topic that matches the unvalued/uninterpretable [Adrs] on Top° and *pro*).
b. Features trigger *Move* (e.g., i) the value(s) of the topic's features are *copied* onto *pro*, and ii) V raises to T° (cf. Roberts 2010b).
c. Features trigger *Agree* (e.g., the value(s) of the features of the goal (i.e. the topic/*pro*) match those of the probe (Top°/T°).

Bearing all this in mind, let us now address the question imposed in Section 2.2.1, i.e., where do T's features come from? Recently, Chomsky (2005, 2008) proposes *Feature Inheritance* as a syntactic notion that signals that the locus of features is C and not T. He emphasizes that φ-features and tense are not an inherent property of T, but rather they belong to C as a phase head. Chomsky proposes that features are inherent only to phase heads, C° and *v**°, and that these features are inherited by T in the (narrow) syntax, i.e., when T is selected by C in the course of derivation. What we would like to stress here is that T in imperatives (and all clause-types) depends on C's features; it cannot and must not probe by itself independently of C (cf. Richards 2012).

In Section 3.1, we have concluded that C in Arabic has φ–, tense and Case features, and thus under *Feature Inheritance* adopted here, T inherits these features from C in the syntax.[26] YA provides empirical evidence that T inherits these features from C, as illustrated in the following examples.

(28) a. ʔadri ʔinna-**k** walay-**k**
I.know that-**2MS** went-**2MS**
'I know that you have gone.'

b. ʔadri ʔinni-**š** walay-**š**

[26] YA does not have Case marking inflections, but Standard Arabic provides independent evidence that C has a Case feature. This is illustrated in (i).
(i) a. ʔallah-u yaʕlamu l-ħaal-a
God-NOM knows the-situation-ACC
'God knows the situation.'
b. ʔinna ʔallah-a yaʕlamu l-ħaal-a
C God-ACC knows the-situation-ACC
'Indeed, God knows the situation.'
In (ia), the topic *ʔallah* appears with a default nominative Case; however, in (ib) it appears with an accusative Case assigned by the C *ʔinna*.

I.know that-**2FS** went-**2FS**
'I know that you have gone.

c. ʔadri ʔinna-**kum** walay-**kum**
I.know that-**2MPL** went-**2MPL**
'I know that you have gone.'

d. ʔadri ʔinni-**kin** walay-**kin**
I.know that-**2FPL** went-**2FPL**
'I know that you have gone.'

These examples very clearly show that T inherits φ-features from C. As is very clear from the glosses in (28a-d), C, i.e., *ʔinna,* shows agreement with the verb in all φ-features (see also Shormani 2017, Shormani and Qarabesh 2018, for discussions). It occurs as *ʔinna-k, ʔinni-š, ʔinna-kum, ʔinni-kin*, where C agrees with the verb in all φ-features. C in Arabic attracts these clitics because it is "a head endowed with φ-features [that] can attract a clitic" (Rouveret 2008: 190).

There are three points to mention here: i) given (27c) the analysis developed here accounts for the licensing of *pro* in imperative structures, in that the unvalued features of T license *pro* under *Agree*, ii) there seem to be A'-chains established between the topic-NP, Top, T and *pro*, and consequently, iii) the interpretation of *pro* depends by and large on the *Feature Matching* taking place between the elements of the established A'-chain. This also suggests that *pro* in Spec,*v*P is coreferentially "linked" with the topic in Spec,TopP.

**5. Imperatives: syntax-discourse interface**

Recall that in NSLs *pro* is referential in nature. However, consider (10) repeated here as (29), for convenience.

(29) a. ʔiktub *pro*!
write.2MS

b. ʔiktub-i *pro*!
write-2FS

c. ʔiktub-uu *pro*!
write-2MPL

d. ʔiktub-ayn *pro*!
write-2FPL

Based on the inflection YA exhibits, several interpretations of *pro* in (29) are enforced. For instance, *pro* can be interpreted as a singular masculine 'you' in (29a), a singular feminine 'you' in (29b), a plural masculine 'you' in (29c), and a singular feminine 'you' in (29d). The enforced interpretation in each can be ascribed to the syntax, as manifested by the agreement inflection attached to the imperative verb, but this interpretation seems to be only "partial". In other words, the "interpretation enforced" by agreement inflection is not full; it is difficult to identify the *actual* referent(s), i.e., the people, *pro* refers to. Suppose these examples are said out of discourse, as is the case of (29), it seems difficult to identify the one/people "functioning" as the addressee(s) in all these examples. For instance, in (29a), it is not at all clear whether the addressee is a 'student', 'audience', 'clerk in an office', etc. That said, if (29a-d) is said without taking into account the topic as the antecedent of *pro*, the interpretation of *pro* will be 'vague' (cf. also Chomsky 1982, Huang 1989, Kayne 2002).

This 'vagueness' seems to have cross-linguistic evidence. Consider the French and English examples in (30a) and (30b), respectively (cf. also Kayne 2002).[27]

(30) a. *pro* écoutes-moi!
listen to me!

b. *pro* write this!

The 'vagueness' *pro* may exhibit in languages like English, more than in Arabic, ensues from the fact that they are "very poor" in agreement inflection, specifically in imperative clauses.[28] However, this 'vagueness' disappears if a topic is mentioned in the imperative sentence.

(31) a. Ali *pro* write!

b. Girls, *pro* open your books!

*pro* in these examples refers to *Ali* in (31a) and *girls* in (31b); and therefore, the interpretation of *pro* is ultimate/full.[29] As far as Arabic is concerned, consider the imperative structures in (32) and (33).

(32) a. ʕali, taʕaal *pro*!
Ali, come.2MS

b. ʕalia, taʕaal-i *pro*!
Alia, come.2FS

---

[27] Kayne (2002: 137) argues that context/discourse plays a crucial role in the interpretation of pronouns in languages like English. He regards examples like (i) as ungrammatical if they are said out of discourse, because their referents do not exist in the conversation world.
(i) a. He is a genius.
b. Watch out! He's got a knife.
As for cases like (ia), Kayne proposes a *silent* topic in the discourse as an antecedent of the pronoun 'He'. Furthermore, Kayne suggests that (ib) has a reading akin to (ii).
(ii) Watch out! That man, he's got a knife.
Kayne (2002) concludes that the interpretation of a pronoun is discourse-/context-bound. Along these lines, Corver (2008:71) stresses that a (null) pronoun may not be adequately and fully interpreted out of "the situational context" (see also Trecci 2006, for Italian).

[28] In fact, French has a verbal inflection that distinguishes 2 person singular from 2 person plural imperatives. This is illustrated in (i).
(i) a. écoute-s-moi
listen-2S-me
'Listen to me!'
b. écoute-z-moi
listen-2PL-me
'Listen to me!'
However, the inflection *–s/-z* seems to have a "limited role" to play in the interpretation of *pro*. This is particularly because French distinguishes person, number and gender. If person and number are clear in (i), gender remains vague/ambiguous. French distinguishes feminine from masculine. Therefore, it will be difficult to identify the addressee in (i). For example, in (ia) it is difficult to identify whether the addressee is a 2 person singular feminine or 2 person singular masculine. And more importantly, if (ia) is said out of context/discourse, it will be difficult to identify whether the addressee is a man or a woman, a female student or a male student, a female clerk, or a male clerk, a female employee or a male employee, and so on. This explicitly suggests that the vagueness of *pro* still persists, even more than that found in YA examples in (29).

(33) a. ƭullaab, taʕaal-uu *pro*!
students, come.2MPL

b. ƭaalibaat, taʕaal-ayn *pro*!
students, come.2FPL

The clear interpretation of *pro*s in (32) and (33) is solely due to the presence of the topics in the C-domain (i.e., Spec,TopP), namely *ʕali* and *ʕaliya* in (32a,b), and *ƭullaab* and *ƭaalibaat* in (33a,b) respectively. Each of these NPs functions as an addressee in the discourse, and it is clear that *pro*'s (full) interpretation depends by and large on these addressees (cf. Moon 2001, Pesetsky and Torrego 2007, Zanuttini et al. 2012).[30]

Based on these facts, we propose that *pro* in imperative structures enters the derivation with valued, but uninterpretable features (cf. Pesetsky and Torrego 2007, Shormani 2017).[31] In the (narrow) syntax, *pro*'s valued features value T's unvalued corresponding ones. This valuation of (T's features) in the syntax, we claim, is not sufficient for *pro's* "Full Interpretation" (because *pro*'s referent is not stated in the world/discourse). However, when the discourse (represented by the topic expression in the CP) comes to play, *pro* obtains its ultimate interpretation. This gives us enough room to postulate that there is some kind of "correlation" between the syntax and discourse. There arises a question in this juncture, however: how does this correlation take place? Recall that *pro* in imperatives is (always) bound by an NP in the C-domain. It follows that this "binding" can take the form of Principle B of Binding Theory (see e.g., Chomsky 1982, 1986, Rizzi 1982, 1986, Jaeggli and Safir 1989). Nevertheless, this account was actually not unproblematic in minimalism, and many problems have already been discussed in the literature, leading to a number of modifications of Binding Theory (see e.g., Reinhart and Reuland 1993, Zwart 2002, Kayne 2002, Hicks 2009, H. Hasegawa 2005, Nunes 2009, Antonenko 2012, Landau 2013). These authors try to modify the 'Binding Principles', and propose various mechanisms. As far as other clause-types are concerned, the problem with these proposals is that there are several cases which they fail to account for (see Antonenko 2012, for details).

Therefore, we propose that "binding" takes the form of coreferentiality and this coreferentiality should be handled in terms of *Agree* as *Match*. Bearing (25) and (27) in mind, we propose (34) as an *Agree* (*Match*) mechanism (cf. Chomsky 2001: 5, Roberts 2010a: 61; Shormani 2017: 151).

[30] This interpretation dependency could be taken as indicative evidence of the relation between the syntax and discourse. It is, in other words, expected that the interpretation of *pro* is restricted to/dependent on the addressee. Along these lines, Zanuttini et al. (2012: 1233) hold "that the subjects of imperative clauses have a restricted interpretation: they refer to, quantify over, or overlap in reference with, the addressee(s)". If we take discourse as a neighboring system of the syntax in the sense of Pesetsky & Torrego (2007: 265), and since the addressee belongs to the discourse of imperatives, as we have argued for, then, we expect that these two systems are coreferentially "correlated" at the interface.

[31] Pesetsky & Torrego (2007: 266) propose that the interpretability of features is an independent operation of their valuation. Their proposal argues against Chomsky's (2001: 5) Valuation/Interpretability Biconditional, which is stated in (i)

(i)Valuation/Interpretability Biconditional
A feature F is uninterpretable iff F is unvalued.

Pesetsky & Torrego provide empirical evidence from English, Latin, Russian, etc. that Chomsky's correlation between valuation and interpretability is not unproblematic. The data discussed in this section add support to Pesetsky & Torrego's postulations.

(34) **Agree**

*Agree* is a matching operation whereby the values of the valued features of *α* (the goal) are copied onto the unvalued feature counterparts of *β* (the probe). Thus, if α's feature matrix contains [$Att_i$:__] and β's contains [$Att_i$: $val_j$], for some feature F = [$Att_i$: ($val_{\{..k..\}}$)], the value $val_k$ should be copied into __ in α's feature matrix (cf. Chomsky 2001: 5, Roberts 2010a: 61).[32]

Cyclicity maintained, given (27) and (34), after *pro* and T merge, there will result a variable matching established between T and *pro*. Thus, if T has the value [Att_α_] for a feature F, then, *pro* will get that value, as a result of *Agree*. It follows that when the topic is merged, it matches and values (and interprets) T's and *pro*'s features (cf. also Sigurðsson 2010, Sigurðsson and Maling 2010).[33] And given the antecedent nature of [Adrs] feature, it is likely that *pro* obtains the feature specifications of the topic before/during *Transfer* to the interfaces (cf. Chomsky 2004, 2008). This story seems to result in local A'-chains. In other words, given that the topic is hosted in the C-domain, and that *pro* is in the T-domain, coreferentiality between the topic, Top°, T° and *pro* results in a local A'-chain.

In local A`-chains, the chain is formed between the topic, Top°, T° and *pro.* A *Match* (*Agree*) relation is then established, whereby Top°'s, T's and *pro*'s unvalued features are valued by the topic (in Spec,TopP). Then, the interpretation of *pro* comes to play, in that *pro* gets the featural specifications of the topic.

In this type of chain, *Agree* takes place as follows. Let an unvalued feature F have the value *u*α, then, *v*α is its valued counterpart. Also, let *u*α be the φ-features (including [Adrs]), of Top°, T° and *pro,* then, when the topic with the value *v*α is merged, an A'-chain is formed between these four elements via a matching *Agree*, thus all unvalued/uninterpretable features get valued/interpreted and deleted at LF. Given (25), (27) and (34), each of these element will get the value [Att:_*v*α_]. This is further schematized in (35).

(35)

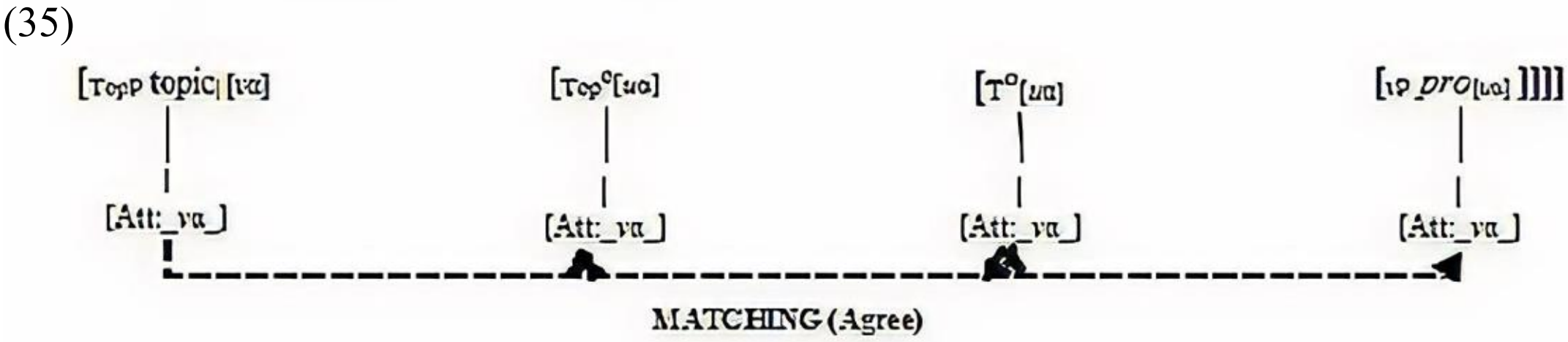

In (35), the coreferentiality between the topic and *pro* takes place locally, i.e., the topic and *pro* are in the same clause. However, the existence of imperative structures like (36) (cf. also (1d)), which are instances of coordinated imperatives, may run counter to our proposal.

[32] In this *attribute–value* mechanism, a valued feature F has the valued pair [Att: val], and the unvalued pair will be [Att: ____]. In our system, we apply this mechanism to characterize the attribute-valuation operation established between a P and a G, as a result of a matching *Agree*. Take T's and *pro*'s nominative Case as an example; T's valued Case feature will be [Case: _nom_] and *pro*'s unvalued Case feature will be [Case:___]. As a result of *Agree*, *pro*'s Case feature will become [Case: _nom_] (cf. also Shormani 2017).

[33] For example, Sigurðsson & Maling (2010: 66-67) argue that "pronouns, overt or silent, are not input to the syntactic computation but its output, that is, syntax computes or 'produces' pronouns by matching and bundling up features…[and that pronoun] arguments are matched against the above mentioned silent logophoric agent ('speaker') and the logophoric patient ('hearer') features in the CP domain."

(36) ʕali$_i$, *pro*1$_i$ biz l-masaaħa wa ʔimsaħ *pro*2$_i$ ṣ-ṣabuura
Ali, take the-duster and erase the-blackboard

‘Ali, take the duster and erase the blackboard!’

Imperative structures like (36) seem to have what we call ‘nonlocal coreferentiality’, a coreferentiality taking place between constituents in nonlocal domains. Put differently, in (36) there is a nonlocal A’-chain resulting from coordination. The two imperative structures, namely *biz pro l-masaaħa* and *ʔimsaħ pro ṣ-ṣabuura* are coordinated, hence resulting in a nonlocal A’-chain. In the whole structure, there are two *pro*s, and only one topic, and thus, the topic *ʕali* functions as an antecedent of both *pro*s. (36) will have the LF representation in (37).

(37)

ʕali$_i$, *pro*1$_i$ biz l-masaaħa wa [<*ʕali*$_i$>] ʔimsaħ *pro*2$_i$ ṣ-ṣabuura
Ali, take the-duster and erase the-blackboard

The second topic in the coordination structure, viz. [<*ʕali*>] is silent and must be so. If it is spelled out; the structure is ungrammatical, as (38) shows (cf. (1d)):

(38) *ʕali$_i$, *pro*1$_i$ biz l-masaaħa wa ʕali$_i$ ʔimsaħ *pro*2$_i$ ṣ-ṣabuura
Ali, take the-duster and Ali erase the-blackboard

In (38), *pro*1 is coreferentially linked with the matrix topic *ʕali*, and *pro*2 is coreferentially linked with the embedded topic *ʕali*, but the embedded topic is itself the same instance of the matrix topic, i.e., the first spelled-out topic. Thus, given Grice’s maxim of quantity, (38) is ruled out. In other words, it is the discourse that rules out (38), which is the core of the proposal pursued here.

Note that imperative structures can also be more complicated than (36), as illustrated in (39) (cf. also 1e).

(39) ʕali$_i$, biz *pro*1$_i$ l-masaaħa wa ʔimsaħ *pro*2$_i$ ṣ-ṣabuurah
Ali take the-duster and erase the-blackboard
wa ʕaliya$_k$ haat-i *pro*3$_k$ l-qalam
and Alia Bring-FS the-pen

‘Ali, take the duster and erase the blackboard, and Alia, bring the marker!’

In (39), there are three *pros*, three T$^o$s, three Top$^o$s, one silent topic, i.e., [<*ʕali*>], and two spelled-out topics, namely *ʕali* and *ʕaliya*. *ʕali* is the antecedent of *pro*1 and *pro*2, while *ʕaliya* is the antecedent of *pro*3. Hence, there are two A’-chains: the first one is nonlocal established between *ʕali*, *pro*1 and *pro*2, and the second is local established between *ʕaliya* and *pro*3. The interpretation of *pro* is determined by the coreferentiality and locality, in that *pro*1 and *pro*2 are interpreted as *ʕali,* while *pro*3 is interpreted as *ʕaliya*.

Note also that (36) can have the form in (40), where the topic is not spelled out.

(40) a. biz *pro* l-masaaħa
take.2MS the-duster
‘Take the duster!’

b. biz *pro*1$_i$ l-masaaħa wa ʔimsaħ *pro*2$_i$ ṣ-ṣabuura

take the-duster and erase the-blackboard

'Take the duster and erase the blackboard!'

Examples like (40a) represent the core imperative structures cross-linguistically, as we have noted so far. Moreover, examples like (40b) are also illustrative of coordination of two distinct imperatives, but without a spelled-out topic. Examples like (36-40), however, call into question two substantial issues: i) what referent can *pro* refer to in (40)?, and, ii) how is this across-sentence coreferentiality licensed in a language ***L***?

Given (27) and (35), we propose (41) as a UG Principle, a principle that licenses *pro*'s (null) topic-antecedents, coreferentiality, and accounts for (non)local A'-chains in structures like (36-40).

**(41) (Silent) Topic-antecedent Principle (STP)**

Discourse respected, in an imperative structure:

i. a topic $\alpha$ may be spelled out; a topic β should be silent if $\alpha = \beta$
ii. $\alpha$ is the antecedent of $pro_{1, 2\ldots n}$, unless β is spelled out
iii. if both $\alpha$ and β are spelled out, then: $[\lambda\alpha\ \lambda\beta\text{: } [pro = \alpha, \text{and } pro = \beta]]$; each within each's domain
iv. a (non)local A\`-chain is formed between α/β and $pro_{1,2\ldots n}$, via *Matching* (*Agree*) relation.

(41) may have a cross-linguistic implementation, which, we think, stems from Grice's Maxim, 'Be concise' (see also Shormani 2017). It could also be assumed that "embedded" topics (i.e., βs) are deleted at LF, presumably in terms of the "Antecedent Contained Deletion" (Fox 2002). Let us now examine the reliability of (41) in YA imperatives. (41i) accounts for structures like (40), in that a preverbal (pro)nominal may be spelled-out, but it does not need to. Consequently, null topics/*pro*s are licensed. (41i) also accounts for (36): when a topic is spelled-out, it cannot be spelled-out once more in the same discourse. (41i) and (41ii) account for the ungrammaticality of structures like (38). (41iii) accounts for structures like (39), i.e., a topic is spelled-out when it is not the same like the first-most spelled-out topic. The former becomes the antecedent of the occurrences of *pro* in its (non)local domain; and therefore, coreferentiality between *pro* and its (antecedent) topic is licensed. *pro* in imperatives seems to be always *lambda* bound by its topic referent in the C-domain (cf. Portner 2007). It has the interpretation of a topic $\alpha$ if it is bound by it, and as a topic β if it is bound by it. Thus, in structures like (36) all instances of *pro*, i.e., *pro*1, *pro*2, are interpreted as *ʕali*. But in (39), while *pro*1 and *pro*2 are interpreted as *ʕali*, *pro*3 is interpreted as *ʕaliya*. A strong piece of evidence supporting this is the agreement between *ʕaliya* and the verb *haat-i*, i.e., both are feminine and singular, while *pro*1 and *pro*2 are interpreted as *ʕali*- they are masculine and singular. Finally, (41iv) accounts for nonlocal A'-chains established between the topmost spelled-out topic, silent topic(s) and more than one instance of *pro* in a given imperative expression.[34]

Another important question worth addressing here is: how does *Agree* work in such structures given the *Phase Impenetrability Condition* (PIC, see Chomsky 2000: 108, 2001: 13)? There must be a mechanism in which *Agree* takes place in such structures, and hence securing violation of the PIC. Reconsidering (36), we find a nonlocal A'-chain established between the first spelled-out topic *ʕali*, Top°1, T°1, *pro*1, [<***ʕali***>], Top°2, T°2 and *pro*2. Note that the latter case involves two TopP-phases and two *v*P-phases. It follows that *Agree* taking place in this context is an AAP relation established *across phases*. Assuming along with Chomsky (2008) and Rouveret (2008) that *Agree* taking place between phases is not subject to PIC effects, these structures give rise to no problem. Chomsky (2008) argues that PIC "holds only for the mappings to the interface, with the effects for narrow syntax automatic" (p.143). He provides empirical evidence that in languages such as Icelandic *Agree* takes place "into a lower phase without intervention in experiencer constructions in which the subject is raised (voiding the intervention effect) and agreement holds with the nominative object of the lower phase" (p. 159). Given this, we propose (42) as an *Agree* mechanism in nonlocal A'-chains, and call it *Agree Across Phases* (AAP).[35]

[34] Note that the reliability of such A'-chains could be argued in terms of chain reduction rules put forth by (Nunes 2004: 21f).

**(42) Agree Across Phases**
In a nonlocal A'-chain,
i. $\alpha_1, \alpha_2, \alpha_3 \ldots \alpha_n$ are phase heads; β is a nonphase head
ii. $\alpha_1$ agrees with $\alpha_2, \alpha_3 \ldots \alpha_n$ in all the relevant features
iii. all α's *u*Fs are valued by α's *v*Fs via *Agree* as *Match*
iv. β is not an intervener since it is a nonphase head

In (42), once $\alpha_1$ is valued for a feature F, it will enter an *Agree* (Match) relation with all the lower instances of α, viz. $\alpha_1, \alpha_2, \alpha_3 \ldots \alpha_n$. In our context, $\alpha_1, \alpha_2, \alpha_3 \ldots \alpha_n$ stand for the instances of Top°s and *v*°s; let F be [Adrs] + φ. Let also F have two values: *u*F and *v*F. If we assign *u*F to every phase head, viz. Top°s or *v*°s, it follows that once the topic is merged, all the instances of *u*F will be valued via *Agree* as roughly schematized in (43).[36]

(43)

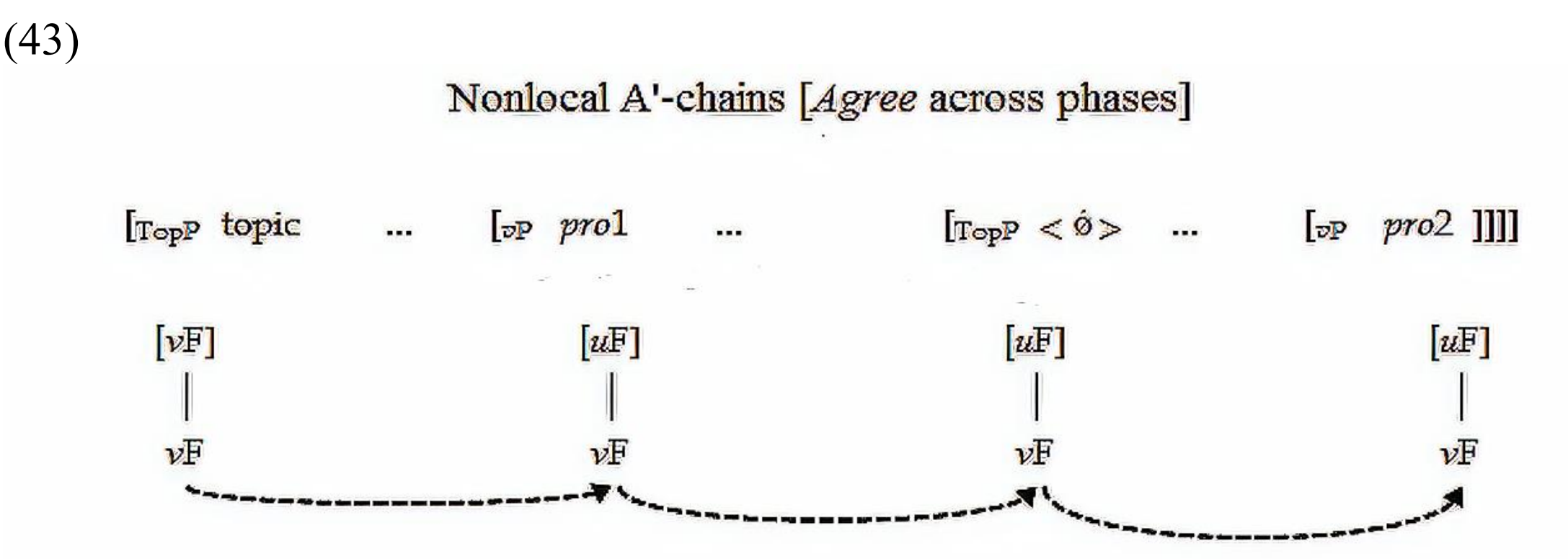

In a cyclic fashion, once a nonlocal A'-chain is established, and once the topmost topic is (re)merged in the specifier of the topmost TopP, the topmost Top° gets valued for F via *Matching* as *v*F. As a result of this MATCHING mechanism, all Top°s and *v*°s of lower phases get the same value, i.e., *v*F for all features including [Adrs].

In this MATCHING (*Agree*) mechanism, imperative structures like (36-40) are accounted for. In (36), for instance, once the nonlocal A'-chain is construed between *ʕali*, *pro*1, the silent topic <***ʕali***> and *pro*2, and once the topmost topic *ʕali* is merged, the topmost Top°, a phase head, gets valued for [Adrs] and φ features, all lower phases heads, i.e., all instance of Top° and *v*° will get valued and interpreted for all these features.[37]

---

[35] Other works that argue and provide evidence for agreement taking place into lower clauses include (e.g., Bošković 2007, Boeckx 2009, Roberts 2010b). They provide evidence from several and different languages, including Hindi, Icelandic, Tsez, Basque, etc.

[36] Note that in terms of (42), heads like T° (i.e. βs) do not have intervention effects because they are nonphase heads (cf. Chomsky 2000, *et seq*). For example, Chomsky's (2000: 123) postulates that T°, as a nonphase head, may not be a defective intervener; and thus, would not block probing from a higher phase into a lower one.

[37] Note that each structure in (36-40) can be longer than its present form, given the recursive nature of imperatives (and other clause-types in general). Take as an example (39), as the most complicated imperative structure involved in this context. There are three instances of *pro,* two nonlocal A'-chains, two spelled out topics, and

## 6. Conclusions and implications

This article answers some questions that have been left open in the existing studies on imperatives. In particular, it answers the questions: i) what are the (pro)nominal constituents showing up in the preverbal position? The article suggests that these (pro)nominal expressions are topics, while, i.e., the thematic subject of imperatives is a 2 *pro* in YA (and cross-linguistically, though the cross-linguistic data involved in this paper may not be sufficient), ii) is discourse involved in licensing and interpreting the imperative null subjects? The article proposes that discourse *is* very much involved, not only in licensing *pro* via the head Top°, but also in determining the ultimate interpretation of imperatives via coreferentiality with the topics. The topic is shown to have a performative function and interpretive import, which is performed by the imperative verb.

And iii) how is interface between syntax and discourse manifested in imperative constructions? Given that the topic is 'housed' in the C-domain, i.e., in Spec,TopP, and that *pro* is merged in the T-domain, i.e., in Spec,*v*P, the article suggests that imperative structures require an analysis based on a correlation of the syntax and discourse at the interface, concluding that the syntax meets with discourse at the interface. This "correlation" takes the form of coreferentiality between the topic(s) and *pro*(s). The former represents the informational structure, and the latter the propositional one. TopP is argued to be a phase in the C-domain, whose head, i.e., Top° is endowed with an [Adrs] feature, which yields a discourse property, and "links" the informational structure with the propositional one.

As for core imperatives like (44) (cf. 1b), we proposed that a null topic is merged in Spec,TopP.

(44)
ʔimnaʕ *pro* l-kalaam!
prevent.2MS the-speech
'Stop talking!'

This null topic *pro* is, then, coreferentially linked with the thematic subject *pro*, as shown in (45a), and roughly schematized in (45b).

(45) a. ʔimnaʕ *pro* l-kalaam!
prevent.2MS the-speech
'Stop talking!'

b. [$_{\text{TopP}}$ *pro*$_i$ …[$_{v\text{P}}$ *pro*$_i$ [$_v$ ʔimnaʕ [$_{\text{VP}}$ [l-kalaam]]]]

The meaning of the null topic/*pro* is, then, determined by the discourse in which the imperative is said. Suppose (44) is said by a teacher in a classroom, and suppose (44) is preceded by a sentence like *Dear students, today we will talk about imperative constructions in English*. It follows that the interpretation/meaning of the null topic *pro* in (45) is *students*.

The coreferentiality between the topic and *pro* is assumed to take the form of a matching *Agree*. This coreferentiality results in two types of A'-chains, viz. local and nonlocal. The former is established between the topic and *pro* in simple imperative structures, while the latter between the topic, and two or more instances of *pro* in coordinated imperatives.

several Top° and *v*° (i.e. phase heads). Applying the MATCHING (*Agree*) mechanism in (43) *Agree* taking place in (39) is accounted for, without any further ado.

To account for all these imperative properties, STP is proposed as a UG Principle necessitated mainly by the referential and performative interpretive requirement of imperatives. STP is, thus, a discourse-syntax principle, determined solely by interpretive import.

Based on the common properties imperatives share cross-linguistically, the analysis proposed in this article has several implications, the most important of which are: i) the analysis suggests that human languages are virtually *pro*-drop languages, at least concerning 2 *pro*, ii) some parts of the analysis add support to the cross-linguistic studies on imperative structures, iii) it could be extended and applied to imperatives across languages.

And vi) as an A'-antecedent of all silent topics and *pros* in a discourse, the Spell-Out of α in (41) could be thought of as an instance of a late Spell-Out driven solely by discourse cross-linguistically. The idea that STP can be implemented across languages ensues from the fact that silent topics are an across-linguistic phenomenon (see e.g., Huang 1984: 545-549, for Chinese and German, N. Hasegawa 1985: 305ff, for Japanese, Hayes and Lahiri 1991, for Bengali, Reinhart 1981: 54ff, Gilligan 1987, Han 1998, 2001, Moon 2001, and more recently, Kayne 2002, Radford 2009, for English, Trecci 2006, Ackema et al. 2006, Frascarelli and Hinterhölzl 2007, Frascarelli 2007, for Italian, Roberts 2010a, for Finnish and Italian, and Demirdache 1988, Shormani 2015, for Arabic, among other authors and languages). A final remark we would like to make here concerns AAP in (42), i.e., AAP. We propose AAP as a mechanism in which *Agree* takes place in nonlocal A'-chains. In our context, it has been applied, and accounts for *Agree* in nonlocal A'-chains construed in examples like (36, 39), and we hope it could be applied to other contexts, say, declaratives, for instance.
To conclude, we would like to point out in this juncture that there are certain cases of imperatives that have not been tackled in this article. The first case concerns negative imperatives such as (46).

(46)
| laa | t-imnaʕ-š | *pro* | l-kalaam! |
|---|---|---|---|
| not | 2-prevent-neg | | the-speech |

‘Don't stop talking.’

Negation in YA clause structure, in general, and in imperatives in particular, is bipartite, i.e. it is formed by two elements: a negative particle *laa* and a negative suffix *–š*. It is almost similar to negation in French as shown in (47) (see also Han 2001, Rowlett 2014).

(47)
| N'écoutes | pas! |
|---|---|
| neg'listen | neg |

‘Don't listen!’

Investigating negative imperatives in YA is very interesting and challenging. In negative imperatives, there are certain changes the imperative verb undergoes. Comparing (44) to (46), it can be observed that the verb *ʔimnaʕ* in (44) is changed into *t-imnaʕ-š*. The second person prefix *ʔi-* changes into *–t*.[38] The latter must appear attached to the verb in negative imperatives, in addition to the negative suffix *–š*.
Another case that we have not addressed in this article is embedded imperatives. In YA, imperatives can occur in embedded clauses like *qul*-clauses ‘say-clauses’ as illustrated in (48).

qulk lak ʔijzaʕ!

[38] Note that *ʔi-* can be taken as an imperative prefix more than a second person prefix; it occurs only in imperatives. However, there are a good deal of imperative verbs which occur without it such as *qum* ‘stand’, *đaʕ* ‘put’, *qul* ‘say’ *daʕ* ‘let’ and so on. However, in negative imperatives, *-t* must appear even in these verbs. For these reasons, we did not gloss *ʔi-* as a second person prefix throughout the paper.

(48) said.I to.you leave
'I told you to leave.'

A very clear example of embedded imperatives is given in (49) (from Classical Arabic), where embedded imperatives appear in *ʔan*-clauses 'that-clauses'. (Qur'an)[39]

(49) "fa-ʔawħinaa ʔilaa musaa ʔan iđrib bi-ʕaṣaaka l-baħar-a"
then-we.inspired to Moses C hit with-stick.your the-sea-ACC
'Then, we told Moses (by inspiration) that hit the sea with your stick!'

Languages differ in allowing or disallowing embedded imperatives; there are languages like English which do not allow imperative embedding, while languages like Arabic or old Scandinavian do (see also Platzack 2007, for old Scandinavian).

Another aspect that we have not addressed in this article is information questions in imperatives. It is widely held that imperative constructions are incompatible with information questions (but see Stegovec 2017, for data from Slovenian). YA provides strong evidence that information questions can be formed in imperatives but only when discourse is much respected. Compare and contrast (50a) with (50b).

(50) a. *What write!?

b. ʔaiš ʔaktub!?
What write.IM
'What should I write?!'

However, it should be noted that examples of information questions in imperatives like (50b) are contextualized. That is to say, (50b) is not acceptable if it is said out of context, which lends further support to our postulation that there is a syntax discourse interface in the interpretation of imperatives. The context in which (50b) is said is that a speaker, i.e. a teacher is trying to tell a frustrated student at the end of exam time "I hereby advise you to write the answer." The addressee, i.e. the student, however, is thinking that there is no use of writing, because there is no enough time to write what he/she wants. The speaker will reply "Ok, just write what you can; it is better than nothing".

[39] Surat Ashu'raa (Verse (52)).